\documentclass[letterpaper]{article}
\usepackage{xcolor}
\usepackage[preprint]{neurips_2026}
\usepackage{graphicx} 
\usepackage[utf8]{inputenc} 
\usepackage[T1]{fontenc}    
\usepackage{hyperref}       
\usepackage{url}            
\usepackage{booktabs}       
\usepackage{amsfonts}       
\usepackage{nicefrac}       
\usepackage{microtype}      
\usepackage{xcolor}         
\usepackage{tabularx}
\usepackage{array}
\usepackage{multirow}
\usepackage{comment}
\usepackage{graphicx}
\usepackage{makecell}
\usepackage{pgfplots}
\usepackage{subcaption}
\usepackage{amsmath}
\usepackage{amssymb}
\usepackage{booktabs}
\usepackage[inline]{enumitem}
\usepgfplotslibrary{groupplots}
\usetikzlibrary{patterns}
\usepackage{tikz}
\usetikzlibrary{arrows.meta,positioning,shapes.geometric}

\definecolor{fixedorange}{RGB}{220,120,30}
\definecolor{rlred}{RGB}{190,40,40}
\definecolor{agenticblue}{RGB}{30,80,180}

\tikzset{
    noactionbar/.style={
        draw=gray!75,
        fill=gray!35,
        postaction={pattern=dots,pattern color=gray!75}
    },
    fixedbar/.style={
        draw=fixedorange,
        fill=fixedorange!45,
        postaction={pattern=horizontal lines,pattern color=fixedorange}
    },
    rlbar/.style={
        draw=rlred,
        fill=rlred!40,
        postaction={pattern=north east lines,pattern color=rlred}
    },
    agenticbar/.style={
        draw=agenticblue,
        fill=agenticblue!45,
        postaction={pattern=crosshatch,pattern color=agenticblue}
    }
}

\newcommand{\resultcell}[2]{%
    \makecell[c]{#1\\[-3pt]\scriptsize($-#2\%$)}%
}

\newcommand{\bestcell}[1]{%
    \makecell[c]{\textbf{#1}\\[-3pt]\scriptsize($0.0\%$)}%
}

\newcommand{\policykey}[2]{
    \makebox[0.15\columnwidth][c]{
        \begin{tikzpicture}[baseline=-0.5ex]
            \draw[#1] (0,0) rectangle (0.34,0.18);
        \end{tikzpicture}
        \hspace{3pt}#2
    }
}

\title{Toward Self-Adaptive Physical AI: Can LLM Agents Manage Long-Horizon Physical Tasks?}

\author{
  Varun Kaushik$^{1}$\quad Yayun Tan$^{2}$  \quad Xiaofan Yu$^{2}$ \\
  $^{1}$The Harker School \\
  $^{2}$University of California, Merced \\
  \texttt{\{vkaushik2027@gmail.com, 
yayuntan@ucmerced.edu, xiaofanyu@ucmerced.edu\}}
}

\newcommand{\xiaofan}[1]{\textcolor{cyan}{#1}}
\definecolor{yayuncolor}{RGB}{180,60,180}

\begin{document}

\maketitle
\begin{abstract}
Large Language Model (LLM) agents offer a promising path toward autonomously managing long-term physical tasks without human intervention. However, physical tasks require agents to continuously observe the environment, make consequential actions, and remain effective as the environment changes. Existing approaches either require substantial data and retraining, or primarily focus on agents operating in the virtual world. In this work, we explore the feasibility of building a self-adaptive physical AI agent that manages long-term physical tasks in a zero-shot manner and adapts to environmental changes without human intervention. We design a multi-agent framework that integrates planning, tool calling, observation, and verification, and evaluate it on agricultural tasks against reinforcement learning (RL) agents under different weather patterns. Our results show that zero-shot LLM agents can achieve comparable management outcomes to RL agents under the same weather pattern and adapt more effectively than RL when evaluated under a shifted environment, highlighting a promising path toward self-adaptive physical AI agents.
\end{abstract}

\section{Introduction}

The recent rise of AI agents, such as OpenClaw~\citep{roumeliotis2026openclaw}, has opened the possibility of using AI systems to autonomously manage physical-world tasks with minimal or no human intervention. Unlike question answering and coding, where systems such as ChatGPT~\citep{openai2022chatgpt,openai2023gpt4} and Claude Code~\citep{anthropic2026claudecode} have demonstrated strong capabilities, physical tasks require agents to continuously interact with and reason about the physical world over extended periods. In such \emph{long-horizon} tasks, an AI agent must repeatedly observe physical phenomena, make decisions whose consequences are irreversible, and verify whether those decisions achieve the intended outcomes. For example, we envision a future in which a user can ask an AI agent to ``take care of my plants.'' The agent would then formulate management objectives, monitor weather and plant conditions, decide when to perform consequential actions such as irrigation or fertilization, and verify that these actions are appropriate, all without human intervention. Such agents could substantially reduce the repetitive human effort required to manage long-horizon physical tasks. We refer to this class of systems as \emph{physical AI agents}.

Despite their promise, building effective physical AI agents remains challenging. First, conventional approaches often require large amounts of high-quality interaction data, which can be expensive or even impractical to collect in the physical world. For example, previous work has shown that training an effective heating, ventilation, and air conditioning (HVAC) controller using reinforcement learning (RL) can require the equivalent of decades of building-operation data, if not managed well~\citep{an2023clue,xu2025efficient}. Second, even when sufficient training data are available, the resulting controller may become ineffective when the deployment environment differs from the training environment, such as when an HVAC controller trained for one building is deployed in another. More broadly, real-world environments are inherently dynamic because of factors such as changing weather and sensor degradation~\citep{yu2024intelligence}. The ability to adapt to such environmental shifts is therefore a critical requirement for physical AI agents.

Previous work has explored several approaches to these challenges. Cyber-physical systems commonly use methods such as model predictive control~\citep{mayne2000constrained} and RL~\citep{sutton2018reinforcement}, including techniques that improve data efficiency~\citep{an2023clue,xu2025efficient}. However, these approaches often remain sensitive to environmental shifts and require new data or retraining when conditions change.
Recent advances in large language models (LLMs) and agentic frameworks offer an alternative path. Systems such as OpenClaw~\citep{roumeliotis2026openclaw} and Hermes~\citep{nousresearch2026hermes} can interpret high-level goals, invoke tools, and execute multi-step tasks autonomously. However, most existing agents operate in virtual environments and still rely on human guidance or verification~\citep{wang2024surveyAgents,zhou2024webarena,liu2025feedback}. Whether LLM-based agents can autonomously manage long-horizon physical tasks and adapt to environmental changes remains largely unexplored.

In this work, we take an initial step toward self-adaptive physical AI agents by studying long-horizon physical task management without retraining or human intervention. We ask two questions: (1) \emph{Can an agentic framework manage a long-horizon physical process in a zero-shot manner, without prior training or fine-tuning and without human intervention?} (2) \emph{Can such an agent autonomously adapt when the physical environment changes?}

To investigate these questions, we design a multi-agent framework that closes the loop among planning, tool use, observation, and verification. The framework decomposes decision making across specialized agents with constrained roles and information access, reducing opportunities for unsupported reasoning and adding independent verification before actions reach the environment. It further introduces reflection-guided feedback that connects the outcomes of previous actions to future decisions, enabling the system to continuously adapt as environmental conditions evolve. We evaluate the framework using a state-of-the-art crop simulator~\citep{solow2025wofostgymcropsimulatorlearning}, which provides detailed observations of weather, soil moisture, nutrient conditions, and crop growth, while supporting management actions such as irrigation and fertilization. We use crop management as a representative long-horizon physical AI task in which decisions are made repeatedly and their effects accumulate over time. The simulator enables controlled, reproducible evaluation across different environmental conditions and direct comparison with state-of-the-art RL baselines. We focus on simulation in this initial study and leave real-world deployment for future work.

The contributions of this paper are summarized as follows:
\begin{itemize}
\item We design a zero-shot multi-agent framework for autonomously managing long-term physical tasks without task-specific training, fine-tuning, or human intervention.
\item We introduce key agent-design mechanisms for physical AI, combining role separation, constrained information access, and independent verification to mitigate hallucinations, together with reflection-guided feedback to support continuous adaptation to changing physical environments.
\item We instantiate the framework in a reproducible crop-management simulator and evaluate its behavior over simulations spanning up to 241 days. Our results show that our zero-shot agentic framework achieves competitive performance under unchanged conditions, substantially outperforms the state-of-the-art RL agent under environmental shift, and uses only a fraction of the fertilizer and irrigation required by RL.
\end{itemize}

\section{Related Work}
\label{sec:related}

Prior work on agents for physical-world task performance can be broadly divided into two categories: (i) traditional, non-LLM-based agents that rely on classical control or reinforcement learning to manage physical processes, and (ii) LLM agents that leverage language model reasoning to plan and act more effectively in physical settings. 

\subsection{Traditional agents for cyber-physical systems}



Previous research in cyber-physical systems (CPS) has long studied how computational systems monitor and control physical processes through feedback loops~\citep{lee2008cps, rajkumar2010cps}. In building automation and industrial process control, widely used approaches include rule-based control~\citep{marik2011advanced, maddalena2020data}, PID control~\citep{marik2011advanced}, and model predictive control (MPC)~\citep{afram2014hvac, qin2003mpc, oldewurtel2012mpc}. These methods are effective in well-characterized physical systems, but often rely on hand-specified rules or explicit system models~\citep{marik2011advanced, oldewurtel2012mpc, serale2018mpc, maddalena2020data}. Recent work has explored reinforcement learning as a general paradigm for physical control, where agents learn policies through interaction with an environment~\citep{recht2019rl, wang2020rl}. For example, \cite{wei2017drl} trained a deep Q-learning controller in simulations; \cite{chen2018rl} used model-free Q-learning for coordinated thermal and ventilation control; and \cite{azuatalam2020rl} applied RL to whole-building control and demand response. Although RL improves adaptivity over fixed controllers, it remains training-dependent~\citep{wang2020rl, recht2019rl}. In contrast, our work enables physical AI agents to adapt from live environmental feedback, without predefined rules, system models, or policy training.

\subsection{LLM-based agents}

Recent work has shown that large language models can be embedded in agentic loops that combine reasoning, tool use, memory, and feedback-driven revision \citep{yao2023react, shinn2023reflexion, schick2023toolformer}. Most empirical demonstrations of LLM agents remain in digital or simulated environments. Their actions are typically executed within simulated worlds  \cite{park2023generative, wang2024voyager} or software systems \cite{zhou2024webarena, xie2024osworld, yang2024sweagent} rather than the physical world. These settings show that LLM agents can plan and revise behavior from digital feedback, but physical process adaptation differs because feedback is noisy and partial, and trials can be costly in time, materials, or safety \citep{dulacarnold2021challenges, ibarz2021train, seifrid2022autonomous}. Embodied LLM and vision-language agents have begun to address this grounding problem in robotics by connecting high-level language reasoning to perception and low-level control \citep{ichter2023saycan, huang2023inner, driess2023palme, zitkovich2023rt2, huang2023voxposer}. Yet many of these systems depend on robot demonstration or pretrained control policies \citep{oneill2024open}. As a result, these systems provide limited evidence of whether LLM agents can adapt autonomously to changes in an ongoing physical process. Our work studies whether such adaptation can emerge from live environmental feedback without retraining or human intervention.

\section{Problem definition}

Without loss of generality, we focus on an agricultural physical AI scenario as a representative case study, covering the process of growing crops and plants. Agricultural management is a good stress-test for physical AI because it combines the properties that make embodied decision making hard: a long timeframe (full growing seasons span months), irreversible actions (fertilizer or irrigation actions can't be undone), and delayed reward (yield is observed at harvest).

We formally formulate the problem as follows. At each decision step $t$, the agent receives an environmental observation $o_t$, which captures the current state of the crop and its growing conditions, and selects a management action $a_t$, such as an irrigation or fertilization decision. Because the effect of an action depends on both the current environment and previous interventions, decisions should account for the interaction history. We denote the available history as
\begin{equation}
h_t=(o_0,a_0,\ldots,o_{t-1},a_{t-1},o_t),
\qquad
a_t=\pi(h_t),
\end{equation}
where $\pi$ denotes the decision policy. As environmental conditions and crop responses evolve throughout the growing season, the policy must continually reassess new observations and adapt subsequent actions based on their observed outcomes.

We formulate the problem as a multi-objective optimization. The primary objective is to maximize the crop yield $Y$ at physiological maturity, while a secondary objective is to avoid unnecessary interventions and resource consumption. We quantify the cumulative resource use over the growing season as
$
C=\sum_{t=0}^{T-1} c(a_t),
$
where $T$ denotes the time of physiological maturity and $c(a_t)$ represents the resource cost associated with action $a_t$. The mathematical formulation of the problem is:
\begin{equation}
\pi^\star \in \operatorname*{arg\,max}_{\pi} \left( \mathbb{E}_{\pi}[Y] -\mathbb{E}_{\pi}[C] \right),
\end{equation}
where $\lambda \geq 0$ controls the trade-off between crop yield and resource use.
This formulation prioritizes successful crop growth and yield while encouraging resource-efficient management whenever additional interventions do not improve the primary outcome.

\section{Physical AI Agent Framework Design}
\label{sec:method}

In this section, we present the design of our Physical AI Agent Framework, a zero-shot prompted framework that performs long-horizon irreversible physical tasks. 
We begin with an overview of the framework (Sec.~\ref{sec:overview}), followed by two key design questions for physical AI agents: how to mitigate hallucinations in closed-loop decision making (Sec.~\ref{sec:hallucination}) and how to continuously adapt to changing environments (Sec.~\ref{sec:adaptation}).

\subsection{Overview}
\label{sec:overview}

\begin{figure}[t]
    \centering
    \includegraphics[width=.9\textwidth]{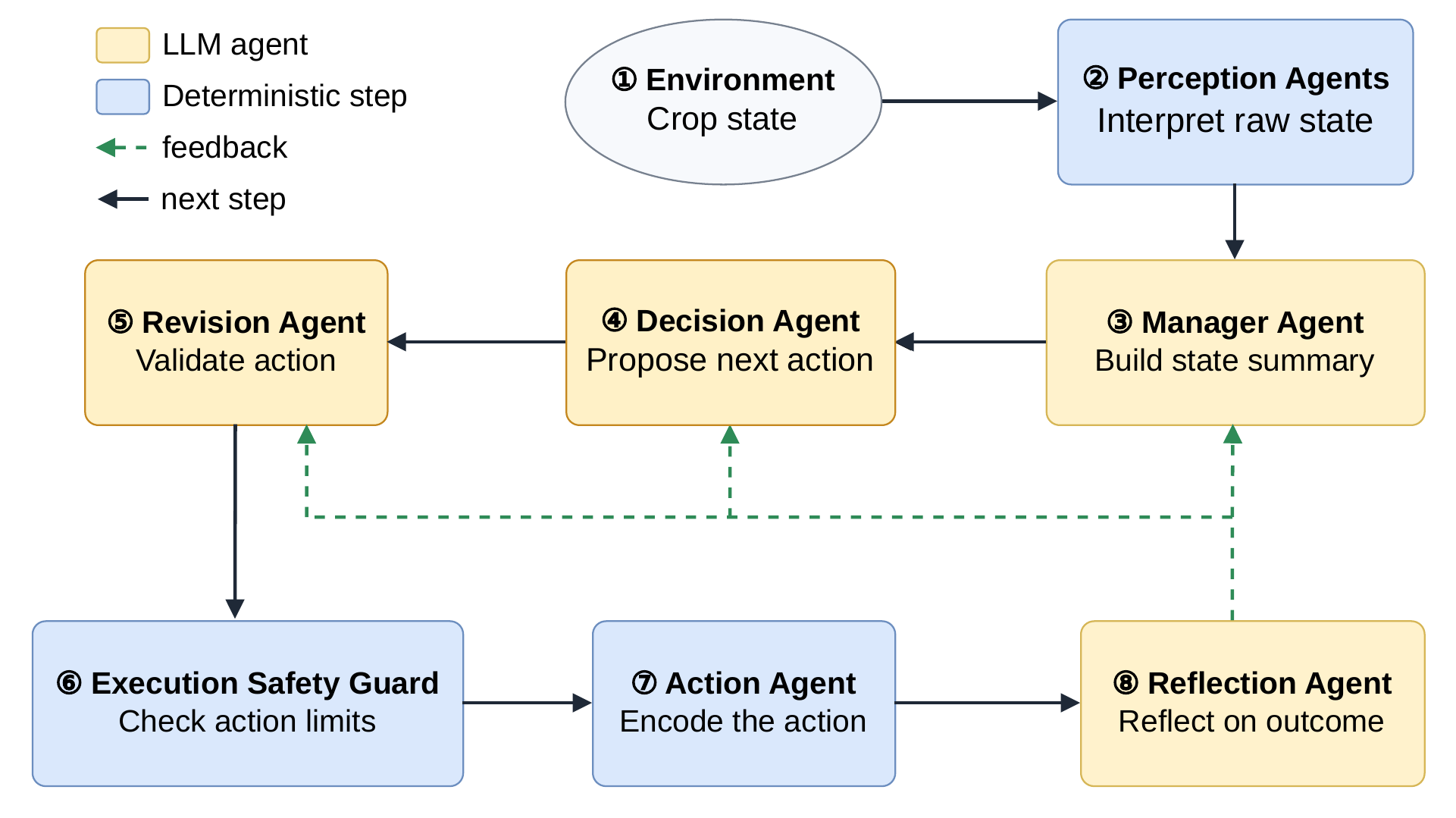}
    \caption{Overview of the proposed Physical AI agent framework, illustrating the closed-loop process of deterministic observation analysis, multi-agent reasoning, action validation and execution, and reflection-guided adaptation. The boxes in yellow represent LLM Agents.}
    \label{fig:framework}
\end{figure}

Figure~\ref{fig:framework} illustrates the overall workflow of our framework. The framework distributes responsibilities across multiple specialized layers and uses daily observations and feedback to refine its decisions throughout the growing period. The process begins with observations of the crop state and growing environment (\textcircled{1}). These observations are interpreted by the \textit{Perception Agents} (\textcircled{2}), which extract relevant information from the raw environmental state and produce structured summaries. The \textit{Manager Agent} (\textcircled{3}) then consolidates these summaries into a coherent representation of the current state. Based on this information, the \textit{Decision Agent} (\textcircled{4}) proposes the next management action, such as irrigation or fertilization. Before execution, the \textit{Revision Agent} (\textcircled{5}) reviews the proposed action and adjusts it when necessary based on the current state and available evidence. The revised action is then checked by the \textit{Execution Safety Guard} (\textcircled{6}) to ensure that it remains within predefined operational limits, after which the \textit{Action Agent} (\textcircled{7}) encodes and executes the action. Finally, the \textit{Reflection Agent} (\textcircled{8}) evaluates the observed outcome of the executed action and feeds its assessment back to the Manager, Decision, and Revision Agents. This feedback closes the decision loop, allowing subsequent decisions to incorporate the outcomes of previous actions and supporting continuous adaptation throughout the growing period.

\subsection{Grounding multi-agent decision making for hallucination mitigation}
\label{sec:hallucination}

Hallucination is a critical challenge when using LLMs for physical-world decision making, as incorrect interpretations or unsupported actions can directly affect the environment. In our setting, hallucinations may include mischaracterizing crop conditions, inferring unsupported deficiencies, or proposing inappropriate interventions. Because many agricultural actions are difficult to reverse, such errors can waste resources or reduce crop yield. 

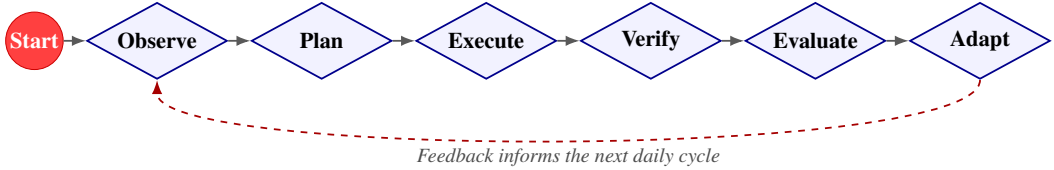
\begin{figure*}[h]
\centering
\resizebox{0.99\textwidth}{!}{%
\begin{tikzpicture}[
    node distance=3.5mm,
    cycleStart/.style={
        circle,
        draw=red!80!black,
        fill=red!75,
        text=white,
        minimum size=9mm,
        inner sep=1pt,
        font=\bfseries
    },
    cycleStage/.style={
        diamond,
        aspect=1.7,
        draw=blue!55!black,
        fill=blue!5,
        line width=0.8pt,
        minimum width=22mm,
        minimum height=12mm,
        inner sep=1pt,
        align=center,
        font=\bfseries
    },
    cycleArrow/.style={
        -{Latex[length=2.2mm]},
        draw=black!65,
        line width=0.8pt
    },
    cycleFeedback/.style={
        -{Latex[length=2.2mm]},
        draw=red!65!black,
        line width=0.8pt,
        dashed
    }
]

\node[cycleStart] (start) {Start};
\node[cycleStage, right=of start] (observe) {Observe};
\node[cycleStage, right=of observe] (plan) {Plan};
\node[cycleStage, right=of plan] (execute) {Execute};
\node[cycleStage, right=of execute] (verify) {Verify};
\node[cycleStage, right=of verify] (evaluate) {Evaluate};
\node[cycleStage, right=of evaluate] (adapt) {Adapt};

\draw[cycleArrow] (start) -- (observe);
\draw[cycleArrow] (observe) -- (plan);
\draw[cycleArrow] (plan) -- (execute);
\draw[cycleArrow] (execute) -- (verify);
\draw[cycleArrow] (verify) -- (evaluate);
\draw[cycleArrow] (evaluate) -- (adapt);

\draw[cycleFeedback]
    (adapt.south)
    .. controls +(0,-1.25) and +(0,-1.25) ..
    node[midway, below, font=\small\itshape, text=black!70]
    {Feedback informs the next daily cycle}
    (observe.south);

\end{tikzpicture}%
}
\caption{The framework's iterative decision cycle.}
\label{fig:decision-cycle}
\end{figure*}

As illustrated in Figure~\ref{fig:decision-cycle}, our framework addresses hallucination throughout the iterative \textit{observe--plan--execute--verify--evaluate--adapt} decision cycle. 
To mitigate hallucinations, we follow two main design principles: (i) separating agent responsibilities and (ii) restricting the information and authority available to each agent. During the \textit{observe} stage, the six Perception Agents are intentionally kept deterministic and limited to their assigned environmental variables. They cannot invent new measurements or freely interpret unrelated information. This helps ensure that the evidence entering the decision cycle remains grounded in the observed environment. During the \textit{plan} stage, the Manager Agent is similarly constrained. Although it uses an LLM, its role is limited to synthesizing the six structured reports, identifying relationships or conflicts among them, and producing a unified state summary. It cannot introduce new measurements or directly select an intervention.

The \textit{execute} and \textit{verify} stages provide an additional layer of protection. The Decision Agent selects an intervention and proposes its amount, while the Revision Agent independently verifies the proposed quantity before execution. To ground this verification, the Revision Agent considers the current observations, the Manager report, permitted action ranges, recent within-episode actions, and previous Reflection feedback. It may approve, increase, or decrease the proposed amount, but it cannot change the intervention type or replace it with a no-op. This separation allows an independently reasoned correction of unsupported or disproportionate quantities without overriding the original choice.

Finally, the Execution Safety Guard provides a deterministic check before the action reaches the environment. It verifies that the revised action is valid, eligible, correctly quantified, and properly encoded for execution, preventing invalid amounts or malformed LLM outputs from being applied. Together, these mechanisms reduce hallucination risk by decomposing the decision process into focused stages, limiting the scope of each agent, and adding independent verification before execution.

\subsection{Reflection-guided adaptation to changing environments}
\label{sec:adaptation}

A physical AI agent must be able to continuously adapt as its environment changes over time. This is especially important in agricultural management, where weather, soil moisture, nutrient availability, and crop requirements can vary substantially throughout the growing season. As a result, decisions that were appropriate under earlier conditions may no longer remain effective. Returning to the decision cycle in Figure~\ref{fig:decision-cycle}, adaptation and feedback therefore play a central role in connecting the outcome of one decision to the next cycle.

\begin{figure}[t]
    \centering
    \includegraphics[width=\textwidth]{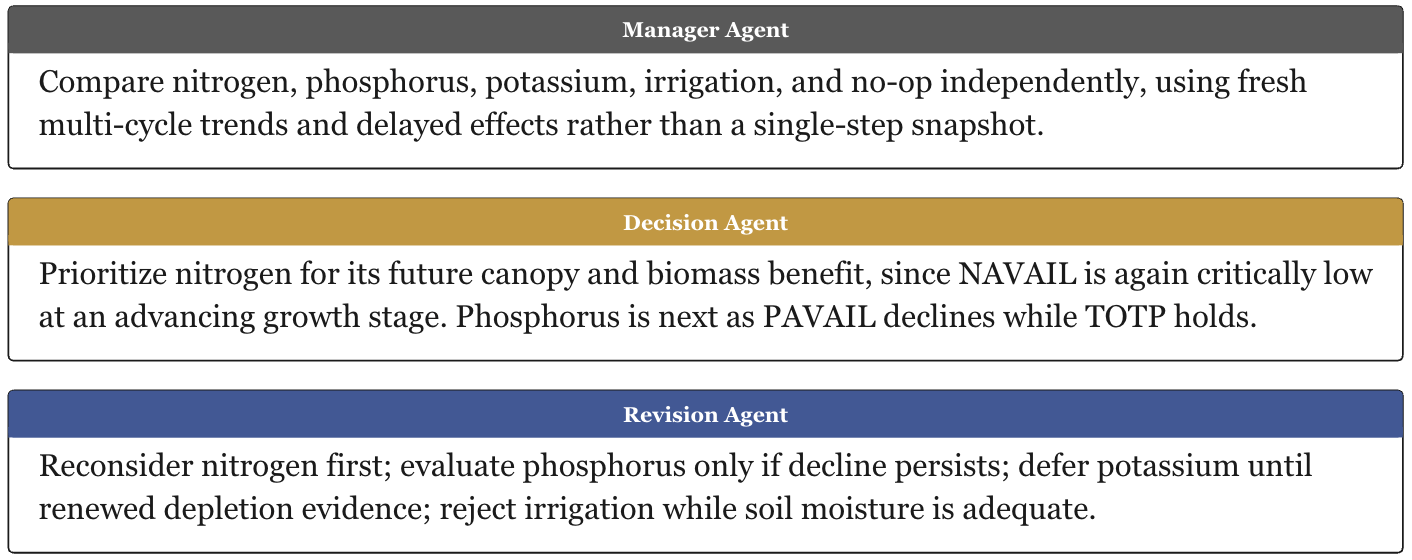}
    \caption{Example of feedback shared by Reflection Agent to other LLM-based Agents.}
    \label{fig:reflection}
\end{figure}

To support this process, our framework uses a Reflection Agent to provide within-episode feedback to the Manager, Decision, and Revision Agents. As illustrated in Figure~\ref{fig:reflection}, this feedback allows the framework to learn from observed outcomes and adjust its reasoning as conditions evolve. Feedback is delivered directly to each agent to avoid losing or distorting important information through intermediate roles.

The feedback is also tailored to each agent's responsibility. The Manager receives guidance on whether its previous interpretation of the environment was supported by subsequent observations and which evidence should be reconsidered. The Decision Agent receives feedback on the effectiveness of the selected intervention and whether alternative choices should be considered in future cycles. The Revision Agent receives feedback focused on the appropriateness of the action quantity. Other components in Figure~\ref{fig:framework} do not receive feedback and remain deterministic, consistent with the design principles described in Section~4.2. Through this role-specific, within-episode feedback mechanism, each subsequent cycle combines new observations with the outcomes of previous actions, allowing the LLM-based agents to update their reasoning and recommendations without implying parameter-level learning or permanent policy changes.

\section{Experiments}

\subsection{Experimental setup}

To implement the framework described in Section~4, we use OpenClaw ~\citep{roumeliotis2026openclaw} to coordinate the multi-agent workflow and manage information exchange between deterministic perception components and LLM-based reasoning agents. The Manager, Decision, Revision, and Reflection Agents are implemented as zero-shot LLM agents using GPT-5.6 Luna~\citep{openai_gpt56_luna_2026}. In total each run made 964 GPT-5.6 calls and used approximately 23.82 million tokens. 

\textbf{Simulator:}
We evaluate all policies in the \texttt{lnpkw-v0} environment of WOFOSTGym, which simulates annual crop growth under limitations in nitrogen, phosphorus, potassium, and water and supports management actions for each resource~\citep{solow2025wofostgymcropsimulatorlearning}. Each episode spans 241 days, corresponding to the simulated wheat growing season.
All evaluated policies receive the same 15 environmental variables and select from the same 17-action space. The observation variables and action space are summarized in Tables~\ref{tab:wofost-observations} and~\ref{tab:wofost-actions}.

We evaluate all policies in two environments to assess their ability to adapt to environmental changes. Environment A uses 2000 weather conditions at latitude $44$, longitude $-99$ (near South Dakota, USA), with \texttt{oregon-3} soil. To evaluate robustness under environmental shift, Environment B changes the weather year, geographic location, and soil type while keeping wheat as the crop. Specifically, it uses 1984 weather conditions at latitude $50$, longitude $5$, with \texttt{oregon-1} soil. This setup represents a realistic distribution shift in which a crop-management policy is deployed under different seasonal, climatic, and soil conditions from those encountered during training.

\begin{table*}[t]
\small
\centering
\caption{The observation variables considered for our agricultural task}
\label{tab:wofost-observations}
\renewcommand{\arraystretch}{1}
\begin{tabularx}{\textwidth}{
    >{\raggedright\arraybackslash}p{2.8cm}
    >{\ttfamily\raggedright\arraybackslash}p{2.0cm}
    >{\raggedright\arraybackslash}X
    >{\raggedright\arraybackslash}p{2.2cm}
}
\toprule
\textbf{Category} &
\multicolumn{1}{>{\raggedright\arraybackslash}p{2.0cm}}{\textbf{Variable}} &
\textbf{Description} &
\textbf{Unit} \\
\midrule
\multirow[c]{4}{2.8cm}{\textbf{Weather}}
    & IRRAD
    & Daily incoming solar radiation
    & J m$^{-2}$ day$^{-1}$ \\
    & TEMP
    & Daily mean air temperature
    & $^\circ$C \\
    & RAIN
    & Daily rainfall
    & cm day$^{-1}$ \\
    & DAYS
    & Number of elapsed simulation days
    & days \\
\midrule
\multirow[c]{3}{2.8cm}{\textbf{Soil and Water}}
    & SM
    & Volumetric soil-moisture content
    & m$^{3}$ m$^{-3}$ \\
    & TOTIRRIG
    & Cumulative irrigation applied during the episode
    & cm \\
\midrule
\multirow[c]{6}{2.8cm}{\textbf{Nutrients}}
    & NAVAIL
    & Plant-available soil nitrogen
    & kg ha$^{-1}$ \\
    & PAVAIL
    & Plant-available soil phosphorus
    & kg ha$^{-1}$ \\
    & KAVAIL
    & Plant-available soil potassium
    & kg ha$^{-1}$ \\
    & TOTN
    & Cumulative nitrogen applied
    & kg ha$^{-1}$ \\
    & TOTP
    & Cumulative phosphorus applied
    & kg ha$^{-1}$ \\
    & TOTK
    & Cumulative potassium applied
    & kg ha$^{-1}$ \\
\midrule
\multirow[c]{3}{2.8cm}{\textbf{Crop Growth}}
    & DVS
    & Crop development stage
    & Dimensionless \\
    & FIN
    & Indicator that growth has finished
    & Binary \\
    & WSO
    & Dry weight of storage organs
    & kg ha$^{-1}$ \\
\bottomrule
\end{tabularx}
\end{table*}

\begin{table}
\centering
\caption{Management actions available}
\label{tab:wofost-actions}
\renewcommand{\arraystretch}{1}
\small

\begin{tabularx}{\columnwidth}{
    >{\raggedright\arraybackslash}p{1.8cm}
    >{\raggedright\arraybackslash}p{1.8cm}
    >{\raggedright\arraybackslash}X
    >{\raggedright\arraybackslash}p{1.4cm}
}
\toprule
\textbf{Action} &
\textbf{Amounts} &
\textbf{Description} &
\textbf{Unit} \\
\midrule

\textbf{No-op}
    & 0
    & Advances the simulation without applying an intervention.
    & -- \\

\midrule

\textbf{Nitrogen}
    & 2, 4, 6, 8
    & Applies nitrogen fertilizer to support plant growth.
    & kg ha$^{-1}$ \\

\midrule

\textbf{Phosphorus}
    & 2, 4, 6, 8
    & Applies phosphorus fertilizer to support root and crop development.
    & kg ha$^{-1}$ \\

\midrule

\textbf{Potassium}
    & 2, 4, 6, 8
    & Applies potassium fertilizer to support crop development and stress response.
    & kg ha$^{-1}$ \\

\midrule

\textbf{Irrigation}
    & 0.5, 1.0, 1.5, 2.0
    & Applies water to increase soil-moisture availability.
    & cm \\

\bottomrule
\end{tabularx}
\end{table}

\textbf{Baselines:}
We compare our framework against three baselines representing unmanaged, fixed, and learned crop-management strategies:
\begin{itemize}
    \item \textit{No-intervention:} Selects no-op throughout the 241-day growing season, providing a control for crop growth without irrigation or fertilization.
    \item \textit{Fixed-intervention:} Applies a predefined management schedule at fixed growth stages. Phosphorus, potassium, and nitrogen are applied at selected stages of crop development, while irrigation is applied at fixed points during tillering, stem elongation, heading, and grain formation.
    \item \textit{Reinforcement learning:} Uses Proximal Policy Optimization (PPO). The PPO model is first trained and evaluated under Environment A. For the environmental-shift experiment, the RL Model is tested under Environment B. Training was performed on a 10-core Apple M2 Pro CPU with 16 GB of unified memory using one environment worker and no distributed computing. The run processed approximately one million timesteps and required 2 hours and 44 minutes. This RL baseline receives these measurements as a normalized numerical vector. 
\end{itemize}
Unlike PPO, our agentic-framework receives structured reports derived exclusively from the same measurements and operates zero-shot without task-specific training or parameter updates.

\textbf{Metrics:}
We use three main metrics to evaluate each policy: (i) cumulative reward, (ii) maximum crop yield, and (iii) total resource use. Together, these metrics capture overall management performance, crop productivity, and resource efficiency.
Cumulative reward measures the total reward accumulated over the full wheat growth cycle according to the WOFOSTGym reward function. Maximum yield is measured by the storage-organ weight (WSO) at physiological maturity and represents the final harvestable crop biomass. Total resource use is computed from the cumulative amounts of nitrogen, phosphorus, potassium, and irrigation applied throughout the episode, providing a measure of the management resources required to achieve the resulting yield.

\begin{figure}
\centering

{\small
\makebox[\columnwidth][c]{
    \policykey{noactionbar}{No action}
    \policykey{fixedbar}{Fixed}
    \policykey{rlbar}{RL}
    \policykey{agenticbar}{Agentic}
}}

\begin{tikzpicture}
\begin{groupplot}[
    group style={
        group size=2 by 1,
        horizontal sep=3mm
    },
    width=0.46\columnwidth,
    height=4.2cm,
    ybar,
    /pgf/bar width=25pt,,
    xmin=0.5,
    xmax=4.5,
    ymin=0,
    ymax=600000,
    ytick= {0,100000,200000,300000,400000,500000,600000},
    scaled y ticks=false,
    xtick=\empty,
    ymajorgrids,
    grid style={gray!20},
    axis background/.style={fill=gray!3},
    axis line style={black!65,line width=0.4pt},
    tick style={black!65},
    nodes near coords={
        \pgfmathprintnumber[
            fixed,
            precision=0,
            1000 sep={}
        ]{\pgfplotspointmeta}
    },
    every node near coord/.append style={
        font=\scriptsize,
        yshift=1pt
    },
    clip=false
]

\nextgroupplot[
    ylabel={Cumulative Reward},
    ylabel style={font=\scriptsize},
    yticklabels={0,100000,200000,300000,400000,500000,600000},
    yticklabel style={font=\scriptsize},
    xlabel={\textbf{(a) Same-environment comparison}},
    xlabel style={
        at={(axis description cs:0.5,+0.08)},
        anchor=north,
        font=\scriptsize
    }
]

\addplot[noactionbar,bar shift=0pt]
coordinates {(1, 27914)};

\addplot[fixedbar,bar shift=0pt]
coordinates {(2,48747)};

\addplot[rlbar,bar shift=0pt]
coordinates {(3,340724)};

\addplot[agenticbar,bar shift=0pt]
coordinates {(4,216390)};

\nextgroupplot[
    yticklabels=\empty,
    xlabel={\textbf{(b) Environment-shift comparison}},
    xlabel style={
        at={(axis description cs:0.5,+0.08)},
        anchor=north,
        font=\scriptsize
    }
]

\addplot[noactionbar,bar shift=0pt]
coordinates {(1,102482)};

\addplot[fixedbar,bar shift=0pt]
coordinates {(2,200553)};

\addplot[rlbar,bar shift=0pt]
coordinates {(3,4798)};

\addplot[agenticbar,bar shift=0pt]
coordinates {(4,453956)};

\end{groupplot}
\end{tikzpicture}
\caption{Cumulative reward for same environment (left) versus shifting environmental (right).All policies are evaluated in the same target environment.}
\label{fig:compact-reward-comparison}
\end{figure}
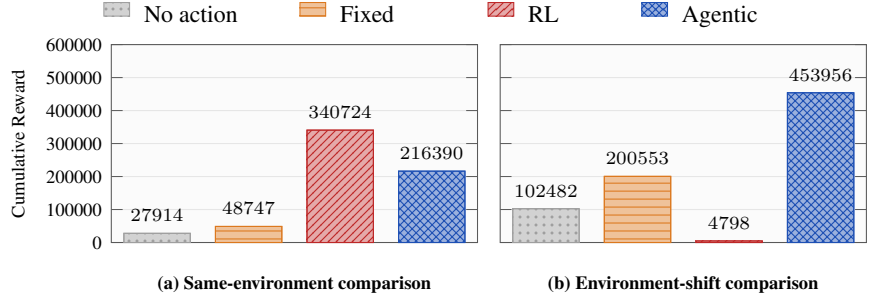

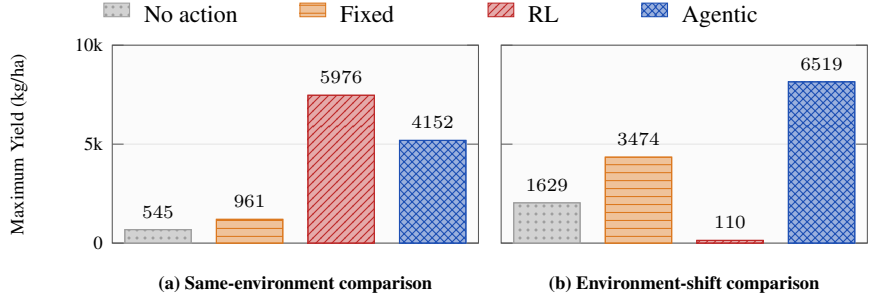
\begin{figure}
\centering

{\small
\makebox[\columnwidth][c]{
    \policykey{noactionbar}{No action}
    \policykey{fixedbar}{Fixed}
    \policykey{rlbar}{RL}
    \policykey{agenticbar}{Agentic}
}}

\begin{tikzpicture}
\begin{groupplot}[
    group style={
        group size=2 by 1,
        horizontal sep=3mm
    },
    width=0.46\columnwidth,
    height=4.2cm,
    ybar,
    /pgf/bar width=25pt,,
    xmin=0.5,
    xmax=4.5,
    ymin=0,
    ymax=8000,
    ytick={0,4000,8000},
    scaled y ticks=false,
    xtick=\empty,
    ymajorgrids,
    grid style={gray!20},
    axis background/.style={fill=gray!3},
    axis line style={black!65,line width=0.4pt},
    tick style={black!65},
    nodes near coords={
        \pgfmathprintnumber[
            fixed,
            precision=0,
            1000 sep={}
        ]{\pgfplotspointmeta}
    },
    every node near coord/.append style={
        font=\scriptsize,
        yshift=1pt
    },
    clip=false
]

\nextgroupplot[
    ylabel={Maximum Yield (kg/ha)},
    ylabel style={font=\scriptsize},
    yticklabels={0,5k,10k},
    yticklabel style={font=\scriptsize},
    xlabel={\textbf{(a) Same-environment comparison}},
    xlabel style={
        at={(axis description cs:0.5,+0.08)},
        anchor=north,
        font=\scriptsize
    }
]

\addplot[noactionbar,bar shift=0pt]
coordinates {(1, 545)};

\addplot[fixedbar,bar shift=0pt]
coordinates {(2,961)};

\addplot[rlbar,bar shift=0pt]
coordinates {(3,5976)};

\addplot[agenticbar,bar shift=0pt]
coordinates {(4,4152)};

\nextgroupplot[
    yticklabels=\empty,
    xlabel={\textbf{(b) Environment-shift comparison}},
    xlabel style={
        at={(axis description cs:0.5,+0.08)},
        anchor=north,
        font=\scriptsize
    }
]

\addplot[noactionbar,bar shift=0pt]
coordinates {(1,1629)};

\addplot[fixedbar,bar shift=0pt]
coordinates {(2,3474)};

\addplot[rlbar,bar shift=0pt]
coordinates {(3,110)};

\addplot[agenticbar,bar shift=0pt]
coordinates {(4,6519)};

\end{groupplot}
\end{tikzpicture}

\caption{Maximum yield for same environment (left) versus shifting environmental (right).All policies are evaluated in the same target environment.}
\label{fig:compact-wso-comparison}
\end{figure}

\subsection{Results}

We first evaluate cumulative reward and maximum yield under two settings: the same environment used to train the RL baseline (Sec.~\ref{sec:res-unchanged}) and a shifted environment with different weather, location, and soil conditions (Sec.~\ref{sec:res-shifted}). We then compare the total resource use of each policy to assess management efficiency (Sec.~\ref{sec:resource-results}).

\subsubsection{Performance under unchanged conditions}
\label{sec:res-unchanged}

Under the unchanged environment, the agentic framework achieves competitive performance with the trained RL policy despite operating zero-shot. As shown in Figure~\ref{fig:compact-reward-comparison} and Figure~\ref{fig:compact-wso-comparison}, the agentic framework achieves a maximum yield of $4,152$ kg/ha compared with $5,976$ kg/ha for RL, corresponding to approximately $69\%$ of the RL yield. Its cumulative reward reaches $216,390$, compared with $340,724$ for RL. These results show that the agentic framework can achieve strong crop-management performance without task-specific training under unchanged environment.

\subsubsection{Performance under environment shift}
\label{sec:res-shifted}

Under environmental shift, the agentic framework substantially outperforms the transferred RL policy. It achieves a cumulative reward of $453,956$, compared with only $4,798$ for RL, and a maximum yield of $6,519$ kg/ha, compared with $110$ kg/ha. As shown in Figure~\ref{fig:compact-reward-comparison} and Figure~\ref{fig:compact-wso-comparison}, the RL policy experiences a severe performance degradation under the new conditions, indicating limited generalization beyond its training environment.

The agentic framework, in contrast, continuously adjusts its decisions based on new observations and reflection feedback. For example, after the Reflection Agent observed that a potassium application increased potassium availability from $0.138$ to $1.284$, the Decision Agent avoided another potassium intervention and instead applied a small amount of nitrogen in response to declining nitrogen availability. This example illustrates how reflection-based feedback helps the framework adjust its management strategy as environmental conditions change.

\subsubsection{Resource usage across environments}
\label{sec:resource-results}

Table~\ref{tab:resource-usage} compares total fertilizer and irrigation use across the two environments. In Environment A, the agentic framework uses $68$/$12$/$72$ kg/ha of N/P/K and $21$ cm of irrigation, compared with $554$/$142$/$258$ kg/ha and $141$ cm for RL. In Environment B, it uses $98$/$16$/$100$ kg/ha and $28.5$ cm of irrigation, compared with $632$/$170$/$360$ kg/ha and $113$ cm for the transferred RL policy.

Across both environments, the agentic framework uses less than $28\%$ of the RL policy's resources in every fertilizer and irrigation category. This substantial reduction shows that its strong performance is not achieved through more aggressive resource application.

\begin{table*}
\centering
\captionof{table}{Resource usage across policies for Environment A (no shift) and Environment B (with shift).}
\label{tab:resource-usage}
\renewcommand{\arraystretch}{1.08}
\setlength{\tabcolsep}{7pt}
\small
\begin{tabular}{@{}llcccc@{}}
\toprule
\textbf{Env} & \textbf{Policy}
& \makecell[c]{\textbf{N}\\\textbf{(kg/ha)}}
& \makecell[c]{\textbf{P}\\\textbf{(kg/ha)}}
& \makecell[c]{\textbf{K}\\\textbf{(kg/ha)}}
& \makecell[c]{\textbf{Irrigation}\\\textbf{(cm)}} \\
\midrule
\multirow{4}{*}{A (no shift)}
& No intervention      & 0   & 0   & 0   & 0.0 \\
& Fixed intervention   & 30  & 6   & 12  & 2.0 \\
& \textcolor{rlred}{RL Model (A on A)} & 554 & 142 & 258 & 141.0 \\
& \textbf{\textcolor{agenticblue}{Agentic framework}}   & 68  & 12  & 72  & 21.0 \\
\midrule
\multirow{4}{*}{B (with shift)}
& No intervention      & 0   & 0   & 0   & 0.0 \\
& Fixed intervention   & 30  & 6   & 12  & 2.0 \\
& \textcolor{rlred}{RL Model (A on B)}  & 632 & 170 & 360 & 113.0 \\
& \textbf{\textcolor{agenticblue}{Agentic framework}} & 98  & 16  & 100 & 28.5 \\
\bottomrule
\end{tabular}
\end{table*}

\section{Limitations}

Although our proposed framework demonstrates promising results, the main limitation of this work is that the agentic-framework was evaluated entirely within the WOFOSTGym crop environment. Although this setting enabled controlled comparisons across environmental conditions, it does not capture complete real-world factors. In future work we plan to evaluate the framework on physical plants using sensors and actuators, allowing us to assess its robustness and practical applicability under real growing conditions.

\section{Conclusion}

We introduced an Agentic AI framework that demonstrates the potential of zero-shot LLM agents to manage long-horizon physical tasks through continuous observation, decision-making, verification, and adaptation. The proposed agentic framework achieved competitive crop performance while using substantially fewer resources than the RL baseline under the original environment. It further maintained strong performance when environmental conditions shifted without retraining, significantly outperforming the RL Model. These findings suggest that structured multi-agent reasoning and outcome-based feedback offer a promising direction for building self-adaptive physical AI systems.

\section*{Acknowledgments}
This research was supported by the Academic Senate Faculty Research Grant and the startup funds at University of California Merced.

\bibliographystyle{plainnat}
\bibliography{references}

@misc{solow2025wofostgymcropsimulatorlearning,
      title={WOFOSTGym: A Crop Simulator for Learning Annual and Perennial Crop Management Strategies}, 
      author={William Solow and Sandhya Saisubramanian and Alan Fern},
      year={2025},
      eprint={2502.19308},
      archivePrefix={arXiv},
      primaryClass={cs.AI},
      url={https://arxiv.org/abs/2502.19308}, 
}

@inproceedings{lee2008cps,
  author    = {Lee, Edward A.},
  title     = {Cyber Physical Systems: Design Challenges},
  booktitle = {2008 11th IEEE International Symposium on Object and Component-Oriented Real-Time Distributed Computing},
  pages     = {363--369},
  year      = {2008},
  doi       = {10.1109/ISORC.2008.25}
}

@inproceedings{rajkumar2010cps,
  author    = {Rajkumar, Ragunathan and Lee, Insup and Sha, Lui and Stankovic, John},
  title     = {Cyber-Physical Systems: The Next Computing Revolution},
  booktitle = {Proceedings of the 47th Design Automation Conference},
  pages     = {731--736},
  year      = {2010},
  doi       = {10.1145/1837274.1837461}
}

@article{marik2011advanced,
  author  = {Marik, K. and Rojicek, J. and Stluka, Petr and Vass, J.},
  title   = {Advanced {HVAC} Control: Theory vs. Reality},
  journal = {IFAC Proceedings Volumes},
  volume  = {44},
  number  = {1},
  pages   = {3108--3113},
  year    = {2011},
  doi     = {10.3182/20110828-6-IT-1002.03085}
}

@article{afram2014hvac,
  author  = {Afram, Abdul and Janabi-Sharifi, Farrokh},
  title   = {Theory and Applications of {HVAC} Control Systems: A Review of Model Predictive Control ({MPC})},
  journal = {Building and Environment},
  volume  = {72},
  pages   = {343--355},
  year    = {2014},
  doi     = {10.1016/j.buildenv.2013.11.016}
}

@article{qin2003mpc,
  author  = {Qin, S. Joe and Badgwell, Thomas A.},
  title   = {A Survey of Industrial Model Predictive Control Technology},
  journal = {Control Engineering Practice},
  volume  = {11},
  number  = {7},
  pages   = {733--764},
  year    = {2003},
  doi     = {10.1016/S0967-0661(02)00186-7}
}

@article{oldewurtel2012mpc,
  author  = {Oldewurtel, Frauke and Parisio, Alessandra and Jones, Colin N. and Gyalistras, Dimitrios and Gwerder, Markus and Stauch, Vanessa and Lehmann, Beat and Morari, Manfred},
  title   = {Use of Model Predictive Control and Weather Forecasts for Energy Efficient Building Climate Control},
  journal = {Energy and Buildings},
  volume  = {45},
  pages   = {15--27},
  year    = {2012},
  doi     = {10.1016/j.enbuild.2011.09.022}
}

@article{serale2018mpc,
  author  = {Serale, Gianluca and Fiorentini, Massimo and Capozzoli, Alfonso and Bernardini, Daniele and Bemporad, Alberto},
  title   = {Model Predictive Control ({MPC}) for Enhancing Building and {HVAC} System Energy Efficiency: Problem Formulation, Applications and Opportunities},
  journal = {Energies},
  volume  = {11},
  number  = {3},
  pages   = {631},
  year    = {2018},
  doi     = {10.3390/en11030631}
}

@article{maddalena2020data,
  author  = {Maddalena, E. T. and Lian, Yingzhao and Jones, Colin N.},
  title   = {Data-Driven Methods for Building Control: A Review and Promising Future Directions},
  journal = {Control Engineering Practice},
  volume  = {95},
  pages   = {104211},
  year    = {2020},
  doi     = {10.1016/j.conengprac.2019.104211}
}

@inproceedings{wei2017drl,
  author    = {Wei, Tianshu and Wang, Yanzhi and Zhu, Qi},
  title     = {Deep Reinforcement Learning for Building {HVAC} Control},
  booktitle = {Proceedings of the 54th Annual Design Automation Conference 2017},
  pages     = {1--6},
  year      = {2017},
  doi       = {10.1145/3061639.3062224}
}

@article{chen2018rl,
  author  = {Chen, Yujiao and Norford, Leslie K. and Samuelson, Holly W. and Malkawi, Ali},
  title   = {Optimal Control of {HVAC} and Window Systems for Natural Ventilation Through Reinforcement Learning},
  journal = {Energy and Buildings},
  volume  = {169},
  pages   = {195--205},
  year    = {2018},
  doi     = {10.1016/j.enbuild.2018.03.051}
}

@article{azuatalam2020rl,
  author  = {Azuatalam, Donald and Lee, Wee-Lih and de Nijs, Frits and Liebman, Ariel},
  title   = {Reinforcement Learning for Whole-Building {HVAC} Control and Demand Response},
  journal = {Energy and AI},
  volume  = {2},
  pages   = {100020},
  year    = {2020},
  doi     = {10.1016/j.egyai.2020.100020}
}

@article{wang2020rl,
  author  = {Wang, Zhe and Hong, Tianzhen},
  title   = {Reinforcement Learning for Building Controls: The Opportunities and Challenges},
  journal = {Applied Energy},
  volume  = {269},
  pages   = {115036},
  year    = {2020},
  doi     = {10.1016/j.apenergy.2020.115036}
}

@article{recht2019rl,
  author  = {Recht, Benjamin},
  title   = {A Tour of Reinforcement Learning: The View from Continuous Control},
  journal = {Annual Review of Control, Robotics, and Autonomous Systems},
  volume  = {2},
  number  = {1},
  pages   = {253--279},
  year    = {2019},
  doi     = {10.1146/annurev-control-053018-023825}
}

@inproceedings{yu2024intelligence,
  title={Lifelong Intelligence Beyond the Edge using Hyperdimensional Computing},
  author={Yu, Xiaofan and Thomas, Anthony and Moreno\textsuperscript{\textdagger}, Ivannia Gomez and Gutierrez\textsuperscript{\textdagger}, Louis and Rosing, Tajana},
  booktitle={Proceedings of the 23rd ACM/IEEE International Conference on Information Processing in Sensor Networks (IPSN)},
  pages={1--13},
  year={2024},
  organization={IEEE},
  url={https://ieeexplore.ieee.org/abstract/document/10577411},
}

@article{xu2025efficient,
  title={Efficient and assured reinforcement learning-based building HVAC control with heterogeneous expert-guided training},
  author={Xu, Shichao and Fu, Yangyang and Wang, Yixuan and Yang, Zhuoran and Huang, Chao and O’Neill, Zheng and Wang, Zhaoran and Zhu, Qi},
  journal={Scientific reports},
  volume={15},
  number={1},
  pages={7677},
  year={2025},
  publisher={Nature Publishing Group UK London}
}

@inproceedings{yao2023react,
  title = {{ReAct}: Synergizing Reasoning and Acting in Language Models},
  author = {Yao, Shunyu and Zhao, Jeffrey and Yu, Dian and Du, Nan and Shafran, Izhak and Narasimhan, Karthik R. and Cao, Yuan},
  booktitle = {International Conference on Learning Representations},
  year = {2023},
  url = {https://openreview.net/forum?id=WE_vluYUL-X}
}

@inproceedings{shinn2023reflexion,
  title = {Reflexion: Language Agents with Verbal Reinforcement Learning},
  author = {Shinn, Noah and Cassano, Federico and Gopinath, Ashwin and Narasimhan, Karthik and Yao, Shunyu},
  booktitle = {Advances in Neural Information Processing Systems},
  volume = {36},
  pages = {8634--8652},
  year = {2023},
  doi = {10.52202/075280-0377},
  url = {https://proceedings.neurips.cc/paper_files/paper/2023/hash/1b44b878bb782e6954cd888628510e90-Abstract-Conference.html}
}

@inproceedings{schick2023toolformer,
  title = {Toolformer: Language Models Can Teach Themselves to Use Tools},
  author = {Schick, Timo and Dwivedi-Yu, Jane and Dess{\`i}, Roberto and Raileanu, Roberta and Lomeli, Maria and Hambro, Eric and Zettlemoyer, Luke and Cancedda, Nicola and Scialom, Thomas},
  booktitle = {Advances in Neural Information Processing Systems},
  volume = {36},
  pages = {68539--68551},
  year = {2023},
  doi = {10.52202/075280-2997},
  url = {https://proceedings.neurips.cc/paper_files/paper/2023/hash/d842425e4bf79ba039352da0f658a906-Abstract-Conference.html}
}

@inproceedings{park2023generative,
  title = {Generative Agents: Interactive Simulacra of Human Behavior},
  author = {Park, Joon Sung and O'Brien, Joseph C. and Cai, Carrie Jun and Morris, Meredith Ringel and Liang, Percy and Bernstein, Michael S.},
  booktitle = {Proceedings of the 36th Annual ACM Symposium on User Interface Software and Technology},
  year = {2023},
  articleno = {2},
  pages = {1--22},
  publisher = {Association for Computing Machinery},
  doi = {10.1145/3586183.3606763}
}

@article{wang2024voyager,
  title = {Voyager: An Open-Ended Embodied Agent with Large Language Models},
  author = {Wang, Guanzhi and Xie, Yuqi and Jiang, Yunfan and Mandlekar, Ajay and Xiao, Chaowei and Zhu, Yuke and Fan, Linxi and Anandkumar, Anima},
  journal = {Transactions on Machine Learning Research},
  year = {2024},
  url = {https://openreview.net/forum?id=ehfRiF0R3a}
}

@inproceedings{xie2024osworld,
  title = {{OSWorld}: Benchmarking Multimodal Agents for Open-Ended Tasks in Real Computer Environments},
  author = {Xie, Tianbao and Zhang, Danyang and Chen, Jixuan and Li, Xiaochuan and Zhao, Siheng and Cao, Ruisheng and Hua, Toh Jing and Zhou, Zhoujun and Shin, Dongchan and Lei, Fangyu and Liu, Yitao and Xu, Yiheng and Zhou, Shuyan and Savarese, Silvio and Xiong, Caiming and Zhong, Victor and Yu, Tao},
  booktitle = {Advances in Neural Information Processing Systems},
  volume = {37},
  year = {2024},
  doi = {10.52202/079017-1650},
  url = {https://proceedings.neurips.cc/paper_files/paper/2024/hash/5d413e48f84dc61244b6be550f1cd8f5-Abstract-Datasets_and_Benchmarks_Track.html}
}

@inproceedings{yang2024sweagent,
  title = {{SWE-agent}: Agent-Computer Interfaces Enable Automated Software Engineering},
  author = {Yang, John and Jimenez, Carlos E. and Wettig, Alexander and Lieret, Kilian and Yao, Shunyu and Narasimhan, Karthik and Press, Ofir},
  booktitle = {Advances in Neural Information Processing Systems},
  volume = {37},
  year = {2024},
  doi = {10.52202/079017-1601},
  url = {https://proceedings.neurips.cc/paper_files/paper/2024/hash/5a7c947568c1b1328ccc5230172e1e7c-Abstract-Conference.html}
}

@inproceedings{ichter2023saycan,
  title = {Do As I Can, Not As I Say: Grounding Language in Robotic Affordances},
  author = {Ichter, Brian and Brohan, Anthony and Chebotar, Yevgen and Finn, Chelsea and Hausman, Karol and Herzog, Alexander and Ho, Daniel and Ibarz, Julian and Irpan, Alex and Jang, Eric and Julian, Ryan and Kalashnikov, Dmitry and Levine, Sergey and Lu, Yao and Parada, Carolina and Rao, Kanishka and Sermanet, Pierre and Toshev, Alexander T. and Vanhoucke, Vincent and Xia, Fei and Xiao, Ted and Xu, Peng and Yan, Mengyuan and Brown, Noah and Ahn, Michael and Cortes, Omar and Sievers, Nicolas and Tan, Clayton and Xu, Sichun and Reyes, Diego and Rettinghouse, Jarek and Quiambao, Jornell and Pastor, Peter and Luu, Linda and Lee, Kuang-Huei and Kuang, Yuheng and Jesmonth, Sally and Joshi, Nikhil J. and Jeffrey, Kyle and Ruano, Rosario Jauregui and Hsu, Jasmine and Gopalakrishnan, Keerthana and David, Byron and Zeng, Andy and Fu, Chuyuan Kelly},
  booktitle = {Proceedings of The 6th Conference on Robot Learning},
  series = {Proceedings of Machine Learning Research},
  volume = {205},
  pages = {287--318},
  year = {2023},
  publisher = {PMLR},
  url = {https://proceedings.mlr.press/v205/ichter23a.html}
}

@inproceedings{huang2023inner,
  title = {Inner Monologue: Embodied Reasoning through Planning with Language Models},
  author = {Huang, Wenlong and Xia, Fei and Xiao, Ted and Chan, Harris and Liang, Jacky and Florence, Pete and Zeng, Andy and Tompson, Jonathan and Mordatch, Igor and Chebotar, Yevgen and Sermanet, Pierre and Jackson, Tomas and Brown, Noah and Luu, Linda and Levine, Sergey and Hausman, Karol and Ichter, Brian},
  booktitle = {Proceedings of The 6th Conference on Robot Learning},
  series = {Proceedings of Machine Learning Research},
  volume = {205},
  pages = {1769--1782},
  year = {2023},
  publisher = {PMLR},
  url = {https://proceedings.mlr.press/v205/huang23c.html}
}

@inproceedings{driess2023palme,
  title = {{PaLM-E}: An Embodied Multimodal Language Model},
  author = {Driess, Danny and Xia, Fei and Sajjadi, Mehdi S. M. and Lynch, Corey and Chowdhery, Aakanksha and Ichter, Brian and Wahid, Ayzaan and Tompson, Jonathan and Vuong, Quan and Yu, Tianhe and Huang, Wenlong and Chebotar, Yevgen and Sermanet, Pierre and Duckworth, Daniel and Levine, Sergey and Vanhoucke, Vincent and Hausman, Karol and Toussaint, Marc and Greff, Klaus and Zeng, Andy and Mordatch, Igor and Florence, Pete},
  booktitle = {International Conference on Machine Learning},
  series = {Proceedings of Machine Learning Research},
  volume = {202},
  pages = {8469--8488},
  year = {2023},
  publisher = {PMLR},
  url = {https://proceedings.mlr.press/v202/driess23a.html}
}

@inproceedings{zitkovich2023rt2,
  title = {{RT-2}: Vision-Language-Action Models Transfer Web Knowledge to Robotic Control},
  author = {Zitkovich, Brianna and Yu, Tianhe and Xu, Sichun and Xu, Peng and Xiao, Ted and Xia, Fei and Wu, Jialin and Wohlhart, Paul and Welker, Stefan and Wahid, Ayzaan and Vuong, Quan and Vanhoucke, Vincent and Tran, Huong and Soricut, Radu and Singh, Anikait and Singh, Jaspiar and Sermanet, Pierre and Sanketi, Pannag R. and Salazar, Grecia and Ryoo, Michael S. and Reymann, Krista and Rao, Kanishka and Pertsch, Karl and Mordatch, Igor and Michalewski, Henryk and Lu, Yao and Levine, Sergey and Lee, Lisa and Lee, Tsang-Wei Edward and Leal, Isabel and Kuang, Yuheng and Kalashnikov, Dmitry and Julian, Ryan and Joshi, Nikhil J. and Irpan, Alex and Ichter, Brian and Hsu, Jasmine and Herzog, Alexander and Hausman, Karol and Gopalakrishnan, Keerthana and Fu, Chuyuan and Florence, Pete and Finn, Chelsea and Dubey, Kumar Avinava and Driess, Danny and Ding, Tianli and Choromanski, Krzysztof Marcin and Chen, Xi and Chebotar, Yevgen and Carbajal, Justice and Brown, Noah and Brohan, Anthony and Arenas, Montserrat Gonzalez and Han, Kehang},
  booktitle = {Proceedings of The 7th Conference on Robot Learning},
  series = {Proceedings of Machine Learning Research},
  volume = {229},
  pages = {2165--2183},
  year = {2023},
  publisher = {PMLR},
  url = {https://proceedings.mlr.press/v229/zitkovich23a.html}
}

@inproceedings{huang2023voxposer,
  title = {{VoxPoser}: Composable 3D Value Maps for Robotic Manipulation with Language Models},
  author = {Huang, Wenlong and Wang, Chen and Zhang, Ruohan and Li, Yunzhu and Wu, Jiajun and Fei-Fei, Li},
  booktitle = {Proceedings of The 7th Conference on Robot Learning},
  series = {Proceedings of Machine Learning Research},
  volume = {229},
  pages = {540--562},
  year = {2023},
  publisher = {PMLR},
  url = {https://proceedings.mlr.press/v229/huang23b.html}
}

@inproceedings{oneill2024open,
  title = {Open X-Embodiment: Robotic Learning Datasets and {RT-X} Models},
  author = {{Open X-Embodiment Collaboration}},
  booktitle = {2024 IEEE International Conference on Robotics and Automation},
  pages = {6892--6903},
  year = {2024},
  doi = {10.1109/ICRA57147.2024.10611477}
}

@article{dulacarnold2021challenges,
  title = {Challenges of Real-World Reinforcement Learning: Definitions, Benchmarks and Analysis},
  author = {Dulac-Arnold, Gabriel and Levine, Nir and Mankowitz, Daniel J. and Li, Jerry and Paduraru, Cosmin and Gowal, Sven and Hester, Todd},
  journal = {Machine Learning},
  volume = {110},
  pages = {2419--2468},
  year = {2021},
  doi = {10.1007/s10994-021-05961-4}
}

@article{ibarz2021train,
  title = {How to Train Your Robot with Deep Reinforcement Learning: Lessons We Have Learned},
  author = {Ibarz, Julian and Tan, Jie and Finn, Chelsea and Kalakrishnan, Mrinal and Pastor, Peter and Levine, Sergey},
  journal = {The International Journal of Robotics Research},
  volume = {40},
  number = {4--5},
  pages = {698--721},
  year = {2021},
  doi = {10.1177/0278364920987859}
}

@article{seifrid2022autonomous,
  title = {Autonomous Chemical Experiments: Challenges and Perspectives on Establishing a Self-Driving Lab},
  author = {Seifrid, Martin and Pollice, Robert and Aguilar-Granda, Andr{\'e}s and Chan, Zamyla Morgan and Hotta, Kazuhiro and Ser, Cher Tian and Vestfrid, Jenya and Wu, Tony C. and Aspuru-Guzik, Al{\'a}n},
  journal = {Accounts of Chemical Research},
  volume = {55},
  number = {17},
  pages = {2454--2466},
  year = {2022},
  doi = {10.1021/acs.accounts.2c00220}
}

@misc{roumeliotis2026openclaw,
  title        = {{OpenClaw and Ollama in Agentic AI: Toward Fully Autonomous and Scalable AI Agent Systems}},
  author       = {Roumeliotis, Konstantinos I. and Sapkota, Ranjan},
  year         = {2026},
  eprint       = {2607.28629},
  archivePrefix = {arXiv},
  primaryClass = {cs.AI},
  doi          = {10.48550/arXiv.2607.28629},
  url          = {https://arxiv.org/abs/2607.28629}
}

@misc{openai2022chatgpt,
  title        = {Introducing ChatGPT},
  author       = {{OpenAI}},
  year         = {2022},
  month        = nov,
  day          = {30},
  url          = {https://openai.com/index/chatgpt/}
}

@article{openai2023gpt4,
  title        = {{GPT-4 Technical Report}},
  author       = {{OpenAI}},
  journal      = {arXiv preprint arXiv:2303.08774},
  year         = {2023},
  doi          = {10.48550/arXiv.2303.08774},
  url          = {https://arxiv.org/abs/2303.08774}
}

@misc{anthropic2026claudecode,
  title        = {{Claude Code: Common Developer Use Cases}},
  author       = {{Anthropic}},
  year         = {2026},
  month        = apr,
  day          = {15},
  url          = {https://support.claude.com/en/articles/14553517-claude-code-common-developer-use-cases}
}

@article{mayne2000constrained,
  title        = {Constrained Model Predictive Control: Stability and Optimality},
  author       = {Mayne, D. Q. and Rawlings, J. B. and Rao, C. V. and Scokaert, P. O. M.},
  journal      = {Automatica},
  volume       = {36},
  number       = {6},
  pages        = {789--814},
  year         = {2000},
  doi          = {10.1016/S0005-1098(99)00214-9}
}

@book{sutton2018reinforcement,
  title        = {Reinforcement Learning: An Introduction},
  author       = {Sutton, Richard S. and Barto, Andrew G.},
  edition      = {2},
  year         = {2018},
  publisher    = {MIT Press},
  address      = {Cambridge, MA},
  isbn         = {9780262039246}
}

@inproceedings{an2023clue,
  author       = {An, Zhiyu and Ding, Xianzhong and Rathee, Arya and Du, Wan},
  title        = {{CLUE: Safe Model-Based RL HVAC ControL Using Epistemic Uncertainty Estimation}},
  booktitle    = {Proceedings of the 10th ACM International Conference on Systems for Energy-Efficient Buildings, Cities, and Transportation},
  pages        = {149--158},
  year         = {2023},
  publisher    = {ACM},
  doi          = {10.1145/3600100.3623742}
}

@misc{nousresearch2026hermes,
  title        = {{Hermes Agent}},
  author       = {{Nous Research}},
  year         = {2026},
  url          = {https://github.com/hermes-agent-org/hermes},
  note         = {Software repository and documentation}
}

@article{wang2024surveyAgents,
  title        = {A Survey on Large Language Model Based Autonomous Agents},
  author       = {Wang, Lei and Ma, Chen and Feng, Xueyang and Zhang, Zeyu and Yang, Hao and Zhang, Jingsen and Chen, Zhiyuan and Tang, Jiakai and Chen, Xu and Lin, Yankai and Zhao, Wayne Xin and Wei, Zhewei and Wen, Jirong},
  journal      = {Frontiers of Computer Science},
  volume       = {18},
  number       = {6},
  pages        = {186345},
  year         = {2024},
  doi          = {10.1007/s11704-024-40231-1}
}

@inproceedings{zhou2024webarena,
  title        = {{WebArena: A Realistic Web Environment for Building Autonomous Agents}},
  author       = {Zhou, Shuyan and Xu, Frank F. and Zhu, Hao and Zhou, Xuhui and Lo, Robert and Sridhar, Abishek and Cheng, Xianyi and Ou, Tianyue and Bisk, Yonatan and Fried, Daniel and Alon, Uri and Neubig, Graham},
  booktitle    = {International Conference on Learning Representations},
  year         = {2024},
  url          = {https://mlanthology.org/iclr/2024/zhou2024iclr-webarena/}
}

@inproceedings{liu2025feedback,
  title        = {A Survey on the Feedback Mechanism of LLM-based AI Agents},
  author       = {Liu, Zhipeng and Bai, Xuefeng and Chen, Kehai and Chen, Xinyang and Li, Xiucheng and Xiang, Yang and Liu, Jin and Li, Hong-Dong and Wang, Yaowei and Nie, Liqiang and Zhang, Min},
  booktitle    = {Proceedings of the Thirty-Fourth International Joint Conference on Artificial Intelligence},
  pages        = {10582--10592},
  year         = {2025},
  doi          = {10.24963/ijcai.2025/1175}
}

@misc{openai_gpt56_luna_2026,
  title        = {{GPT-5.6 Luna}},
  author       = {{OpenAI}},
  year         = {2026},
  url          = {https://developers.openai.com/api/docs/models/gpt-5.6-luna},
}

\end{document}